\documentclass[conference]{IEEEtran}
\IEEEoverridecommandlockouts
\usepackage{cite}
\usepackage{amsmath,amssymb,amsfonts}
\usepackage{algorithmic}
\usepackage{graphicx}
\usepackage{textcomp}
\usepackage{xcolor}
\usepackage{todonotes}
\presetkeys{todonotes}{inline}{}
\usepackage{mathtools}
\usepackage{booktabs}
\def\multiset#1#2{\ensuremath{\left(\kern-.3em\left(\genfrac{}{}{0pt}{}{#1}{#2}\right)\kern-.3em\right)}}
\def\BibTeX{{\rm B\kern-.05em{\sc i\kern-.025em b}\kern-.08em
    T\kern-.1667em\lower.7ex\hbox{E}\kern-.125emX}}

\newcommand{\gameR}{$\mathcal{R}$}

\begin{document}

\title{Rinascimento: using event-value functions for playing Splendor}

 \author{\IEEEauthorblockN{Ivan Bravi}
 \IEEEauthorblockA{
    \textit{School of Electronic Engineering and Computer Science}\\
    \textit{Queen Mary University of London}\\
    London, United Kingdom \\
    i.bravi@qmul.ac.uk
}
\and
\IEEEauthorblockN{Simon M. Lucas}
 \IEEEauthorblockA{
    \textit{School of Electronic Engineering and Computer Science}\\
    \textit{Queen Mary University of London}\\
    London, United Kingdom \\
    simon.lucas@qmul.ac.uk
}}



\maketitle

\begin{abstract}
	In the realm of games research, Artificial General Intelligence algorithms often use score as main reward signal for learning or playing actions. However this has shown its severe limitations when the point rewards are very rare or absent until the end of the game. This paper proposes a new approach based on event logging: the game state triggers an event every time one of its features changes. These events are processed by an Event-value Function (EF) that assigns a value to a single action or a sequence.

	The experiments have shown that such approach can mitigate the problem of scarce point rewards and improve the AI performance. Furthermore this represents a step forward in controlling the strategy adopted by the artificial agent, by describing a much richer and controllable behavioural space through the EF. Tuned EF are able to neatly synthesise the relevance of the events in the game. Agents using an EF show more robust when playing games with several opponents.
\end{abstract}

\begin{IEEEkeywords}
artificial general intelligence, benchmark, game-playing, hyper-parameter optimisation
\end{IEEEkeywords}

\section{Introduction}
	
    
	In the field of Game AI there are many possible applications of AI algorithms: game-playing, procedural content generation, player modelling, analytics and more. Game-playing algorithms are normally based on using a model of the environment to plan their actions (e.g. Monte Carlo Tree Search (MCTS)\cite{browne2012survey}, Rolling Horizon Evolutionary Algorithm (RHEA)\cite{perez2013rolling}), or on using reinforcement learning to learn a policy or value function (e.g. Deep Reinforcement Learning (DRL)\cite{mnih2013playing}). These approaches can also be combined to great effect e.g. AlphaZero \cite{silver2017mastering}.
	

	This learning process usually exploits the presence of a in-game score as a measure of good game-play, a reward signal. Such approach is commonly used in general game-playing frameworks such as the Arcade Learning Environment (ALE) \cite{bellemare2013arcade} and the General Video Game AI (GVGAI)\cite{perez2019general}. Unfortunately these rewards can be very rare or absent until the end of the game, making the use of planning or learning particularly expensive.
	
	In this paper we explore a new idea: to directly learn the value of events.  The intuition behind this, is that events are key to any game, and that it may be easier to learn their value directly, than to learn a state-value function that typically reflects the combined effects of many different events that occurred at different times. The counter-argument is that the game state represents a distillation of all those events, and so contains all that matters. However, in cases where the game state is not fully observable, it may be easier to learn the value of events. Furthermore, even if the information content was theoretically equivalent, learning the values of events directly could still be advantageous.

	Another important point that we try to approach in this paper, and unfortunately rarely approached in Game AI, is guaranteeing diversity in the AI's game-playing style (later in the paper we will use \textit{game-playing style} and \textit{behaviour} interchangeably). For SFP methods we can achieve some diversity by choosing different hyperparameter settings for an agent (e.g. how far ahead it is planning), but we are still bounded to winning or scoring signals. More specifically, we don't have much control on what happens between the signals, e.g. in Super Mario Bros.\texttrademark~ what Mario does between killing a Koopa and the next. A different story is for DRL where the black-box nature of the algorithms makes characterising their ability to express different styles not as straightforward.
	Such diversity is promoted through a set event-based features that could allow for a richer variety of playing styles than a single value signal as the score.
	Behaviour expressivity is very crucial when it comes to automatic playtesting, we want an algorithm/player model that can express enough strategies so to completely cover the space of strategies in the game tested. This becomes even more crucial in a multiplayer game, in fact, opponents will influence the optimal strategy required. For this reasons, in the context of this paper any improved proficiency is welcomed but won't be a make or break factor.

	To summarise, we present a novel approach to reshaping the reward landscape based on \textit{events}: rewards are just the culmination of a series of events triggered by the players and the environment. Monitoring events is a more fine-grained approach that can fill in the gaps between sparse rewards creating a gradient to follow.
	Events are clearly game-specific, however the methodology making use of them can be still regarded as generic.

	Section \ref{sec:background} discusses the state of the art where this research takes place, Section \ref{sec:event} gives a formal definition of the \textit{event-value function} and how that is implemented. Section \ref{sec:experiments} describes the experiments ran which are then discussed in Section \ref{sec:discussion}. Finally Section \ref{sec:conclusion} draws the final conclusion giving a glimpse of possible future work.


\section{Background}\label{sec:background}

	The problem of reward scarcity is central in game AI, it is well known the case of the game Montezuma's Revenge, part of the ALE. This game offers very sparse rewards to the player and only when performing specific actions. Pohlen et al. \cite{pohlen2018observe} managed to improve the state of the art performance of RL algorithms by designing an algorithm based on heavy exploration. Others have reached superhuman performance but only by using demonstration-based approaches by either providing YouTube videos of human players \cite{aytar2018playing} or a single successful demonstration \cite{salimans2018learning}.

	The idea of capturing the dynamics of a game has been explored by taking different approaches, some borrow concepts from psychology, some rephrase the reinforcement learning problem for transfer learning while others model the incentive for exploration as a measure of new dynamics discovered. The following paragraphs provide insight on each one.
	
	Holmgard et al. \cite{holmgard2018automated}, use the concept of \textit{affordance} to describe specific player-styles or behaviours. The concept of affordance is closely related to the question "what can I do with this object" and that's why it is closely related to the fields that make of interaction their focus (e.g. Human Computer Interaction and branches of Robotics). In \cite{holmgard2018automated} the authors associate a metric to each affordance, then design AI agents that plan their actions to maximise/minimise a selection of such metrics in order to show specific behavioural traits. These agents were created in order to develop a portfolio of \textit{personas} to be used for playtesting purposes.

	Perez et al. \cite{perez2014knowledge} have developed a MCTS modification promoting the exploitation of actions that bring the player in game states that trigger new and unseen interactions. This is done by collecting statistics on new interactions during the \textit{rollout} phase of MCTS. Their experiment showed that using information coming from the dynamics of the environment can bring significant performance improvement.

	In the field or Reinforcement Learning the concept of \textit{successor feature} (SF) has been described by \cite{barreto2017successor}. Given the typical definition of Markov Decision Process (MDP) as a tuple of: $S$, set of states; $A$, set of actions; $p$, transition function between states; $r$, reward function. Based on the work of Dayan \cite{dayan1993improving}, the reward function from state $s$ performing action $a$ to state $s'$ can be redefined as $r(s,a,s')=\boldsymbol\phi(s,a,s')^\top \bullet \mathbf{w}$, where $\boldsymbol\phi \in \mathbb{R}$ and $\mathbf{w}$ is a vector of weights. The function $\boldsymbol\phi$ yields a vector synthesising the dynamics of the state transition.
	As a consequence we can derive a new description of the action-value function $q$ applying the new form of $r(s,a,s')$ to its definition: $q(s,a) = \boldsymbol\psi(s,a,s')^\top \bullet \mathbf{w}$. The function $\psi(s,a,s')$ will decouple the dynamics of the environment from the reward function in the shape of \textit{successor features}.


	\subsection{Rinascimento}

		Rinascimento (also stylised \gameR) is a framework\footnotemark~for the development of Game AI, it is based on the popular board game Splendor published by Space Cowboys and designed by Marc Andr\'{e} \cite{bravi2019rinascimento}. Splendor is a turn-based multiplayer board game, the objective is reaching 15 prestige points first, points are obtained by either buying bonus cards purchased using tokens or attracting nobles based on the cards purchased. Four cards are laid face-up from each deck, these can be bought or reserved and bought later, in both cases it moves to the player's hand. When a card is bought the player receives the card's points, then a new card is drawn from the same deck and placed face-up on the table. A card can be reserved if it is face-up or face-down but on top of the deck, in this case it will be revealed to the other players only when purchased. A player can pick either three tokens of different suit or two of the same but only if there are more than 3. Each player has to perform a single action each turn and never own more than 10 tokens.

		The framework can be used to play Splendor-like games, in fact, the parameters in the game's rules are exposed so that they can be easily changed. The same applies to the decks of cards in the game, Splendor has 3 decks but \gameR~can support any number of decks.

		\footnotetext{github.com/ivanbravi/RinascimentoFramework}

		\gameR~implements a Forward Model (FM) that can be used by planning algorithms to simulate future game states given the actions performed by the players. The use of a FM doesn't affect the real game state but it provides "oracle" skills to the agents during their decision making. It has been used to test SFP algorithms \cite{bravi2019rinascimento}, particular attention was given to hyperparameter tuning. The experiments have shown that algorithms using prestige points variations as reward signal, prefer very short action sequences when planning. This is likely due to the highly stochastic nature of the game coming from decks shuffling, partially observable game states and opponents actions.

		Even though \gameR~can potentially express a multitude of Splendor-like games in this paper we will mainly deal with the two-player version of the original game, unless otherwise stated.
		The nature of the game makes enumerating the legal action set computationally expensive (see \cite{bravi2019rinascimento} for a more detailed explanation linked to the parametric nature of the framework). To avoid this enumeration, \gameR~lets the agent to sample randomly the action space with the option of controlling the sampling by setting the random seed. Such randomly generated actions can be illegal if performed in a different game state, that's why extra care needs to be taken when designing the agents.

	\subsection{Algorithms}

		Statistical Forward Planning algorithms are used in games where a Forward Model is available, this will allow the algorithm to simulate actions in the future without affecting the current state of the games. Both Rolling Horizon Evolutionary Algorithm and Monte Carlo Tree Search have shown remarkable performance in games-based competitions such as the GVGAI Planning Competition and the Fighting Game AI Competition \cite{noguchi2019improving}.
		In \cite{bravi2019rinascimento} Bravi et al. have implemented a number of game-playing algorithms:
		\begin{itemize}
			\item RND: random actions;
			\item OSLA: one step look ahead agent;
			\item MCTS: an implementation of MCTS using the Upper Confidence Bound (UCB) formula for the node \textit{selection} and Iterative Widening for dealing with the unknown action space size in the \textit{expansion} phase;
			\item BMRH: Branching Mutation RHEA evolves an action sequence, during the mutation phase a point in the sequence is selected and from there on the actions are mutated with new legal random actions by rolling the state.
			\item SRH: Seeded RHEA evolves an action sequence made of seeds thus circumventing the issue of actions becoming illegal. The seeds are fed to the action random sampling thus fixing the actions generated.
		\end{itemize}
		The hyperparameters of the three different SFP agents have been tuned for optimal performance against the OSLA agent, such agents will be addressed as MCTS*, BMRH* and SRH*.
		In this work we will use the BMRH agent and its hyperparameter space described in \cite{bravi2019rinascimento} for two main reasons: it has proven similar peak performance as MCTS and its hyperparameter space resulted much denser of well performing configurations than MCTS and SRH.

		\subsubsection{Branching Mutation Rolling Horizon}
\begin{figure}[!t]
		    \centering
		    \includegraphics[width=\columnwidth]{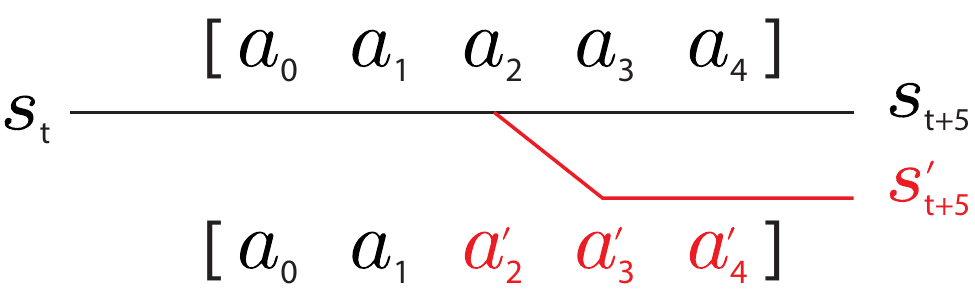}
		    \caption{\label{fig:branching} Branching mutation on a sequence with 5 actions. Actions are copied up to the mutation point $a_2$, the following actions are then picked randomly.}
			\end{figure}

			BMRH is a type Rolling Horizon Evolutionary Algorithm introduced in \cite{bravi2019rinascimento}, for every game tick it evolves sequences of actions evaluating them using a forward model. Such evaluation is based on some value function that measures the quality of the sequence, it is usually the score increment from the starting to final state. Its main feature is the mutation operator: during the creation of a new offspring the original sequence is copied one action at a time and also played until the mutation point, from there on a new action is sampled and added to the sequence until the end of the sequence. An visual representation of the process is shown in Figure \ref{fig:branching}. At the end of the evolution it picks the best sequence and returns the first action.

			The actions in a sequence are dependent to each other in a cascade fashion: an earlier action could make a later action illegal. Given the tight constraints of this game on legal actions (all resources are very different and limited) we implemented the branching mutation in order to reduce the impact of illegal actions to the evolution. For the same reason no crossover operator is used in creating new offsprings.

	\subsection{Hyperparameter Tuning}\label{sec:background:ntbea}

		Most algorithms have several parameters that can be adjusted offline to modify its online execution. For example, the maximum tree depth reached by MCTS and the exploration constant of UCB; examples for RHEA are the mutation probability or the action sequence length. Such parameters are called hyperparameters and in scenarios like game-playing they can have a big impact on the agent's performance.

		There's a vast number of applications of hyperparameter tuning in many academic fields, however in the field of game AI its application is still limited. In \cite{lucas2019efficient}, Lucas et al. have compared several optimisation algorithms in the task of tuning a game-playing RHEA AI and concluded that the N-Tuple Bandit Evolutionary Algorithm (NTBEA) is the best.
		NTBEA, introduced in \cite{lucas2018n}, is a model-based optimisation algorithm, it builds a model of the hyperparameter space using the information gathered by the fitness evaluations of the hyperparameter candidate solutions. This information is stored in multi-armed bandits where each arm of a bandit is a configuration of N specific hyperparameters.

\section{Event-value function} \label{sec:event}

	In a RL scenario, where $S$ is the set of states and $A$ is the set of actions, our main objective would be learning a value function. This can be an state-value function in the shape of $v: S \to \mathbb{R}$ or an action-value function as $q: S \times A \to \mathbb{R}$. However in this work we introduce the concept of event-value function: $h(E_{s \to s'})$ where $s$ is the current state, $s'$ is a future reached playing $\mathbf{a}$ sequence of $n$ actions, $E_{s \to s'}$ is the set of events happening between $s$ and $s'$ as $E$ is the set of all possible events. This function requires a model $m: S \times A \to E$ that generates the events triggered by $a$ from $s$.
	 
	 As a first step in developing this approach we have de-constructed the function $h$ as $h_{\mathbf{w}}(E_{s \to s'}) = f_{\mathbf{w}}(\sigma(E_{s \to s'}))$. The function $\sigma$ synthesises $\boldsymbol\theta \in \mathbb{R}^t$ features from the list of events $E_{s \to s'}$ while $f_{\mathbf{w}}(\boldsymbol\theta)$ is a parameterised \textit{mixer function} in $\mathbf{w} \in \mathbb{R}^t$ weights.

    \subsection{Event Logging} \label{sec:event}
    	
    	In \gameR~ the game state is made up by the following elements: 3 decks of card, face-up cards, nobles, common tokens, joker tokens, 2 player states. Each player state is made up by: points, purchased cards, common tokens, joker tokens, reserved cards, hidden reserved cards.
    	Whenever the engine performs an action that modifies the game state this raises an event which is forwarded to a list of subscribed loggers. An event is described by the fields: \textit{tick}, when it happened; \textit{who}: who triggered it; \textit{type}, unique identifier of the type of event in the range $[0,\#types-1]$; \textit{duration}, how long it lasted; \textit{durationType}, whether the event is instant, delayed or durative; \textit{attributes}, dictionary of attributes characterising the event; \textit{signature}, list of possible attribute keys; \textit{trigger}, what action triggered it.
    	Such description provides very rich information that can be used by the player to make more informed decisions. This definition is general enough to be applied to most games, in fact, it was compiled by referring to several AI game playing competitions. However, in this work we explore the use of just two fields: \textit{who} and \textit{type}.
    	For the specific case of \gameR~we have defined 18 different event types, see Table \ref{table:events}. In particular, when it comes to token related events, an event is triggered for each single token type. The column $\mathbf{Type}^{id}$ assigns a unique id to each event. $\mathbf{Type}^{hc}$, instead, groups the events in 5 hand-crafted macro-events and filters out minor events (-1 ids).

    	\begin{table}[h]
    	\caption{List of all the events, $P_{i}$ is the $i$-th player, E for events triggered by the passive rule of \gameR's engine. When a state element has several events these are listed in the Event column separated by a comma and so are the relative ids.}
		\label{table:events}
		\centering
		\resizebox{\columnwidth}{!}{
		\begin{tabular}{l|l|l|l|l}
		\textbf{State element} & \textbf{Event}         & \textbf{Who} & $\mathbf{Type}^{id}$ & $\mathbf{Type}^{hc}$  \\
		\hline
		Noble                  & place, take, receive     & $P_{i}$          & 7, 0, 14       & -1, -1, 3               \\
		Table's token          & increase, decrease      & $P_{i}$          & 1, 2            & -1, -1                   \\
		Table's joker          & increase, decrease      & $P_{i}$          & 3, 4            & -1, -1                   \\
		Table's card           & draw, place             & $P_{i}$          & 5, 6            & -1, -1                   \\
		Player's token         & increase, decrease      & $P_{i}$          & 8, 9            & 0, -1                    \\
		Player's joker         & increase, decrease      & $P_{i}$          & 10, 11          & 0, -1                    \\
		Table's card           & reserve, hidden         & $P_{i}$          & 13, 12          & 2, 1                     \\
		Player's points        & from card              & $P_{i}$          & 16               & 4                         \\
		Player's points        & from noble             & E                & 17               & 4                         \\
		Player's card          & buy                    & $P_{i}$          & 15               & -1                        \\
		\end{tabular}}
		\end{table}
	
	\subsection{Logging dynamics}
\begin{figure}[!t]
		    \centering
		    \includegraphics[width=\columnwidth]{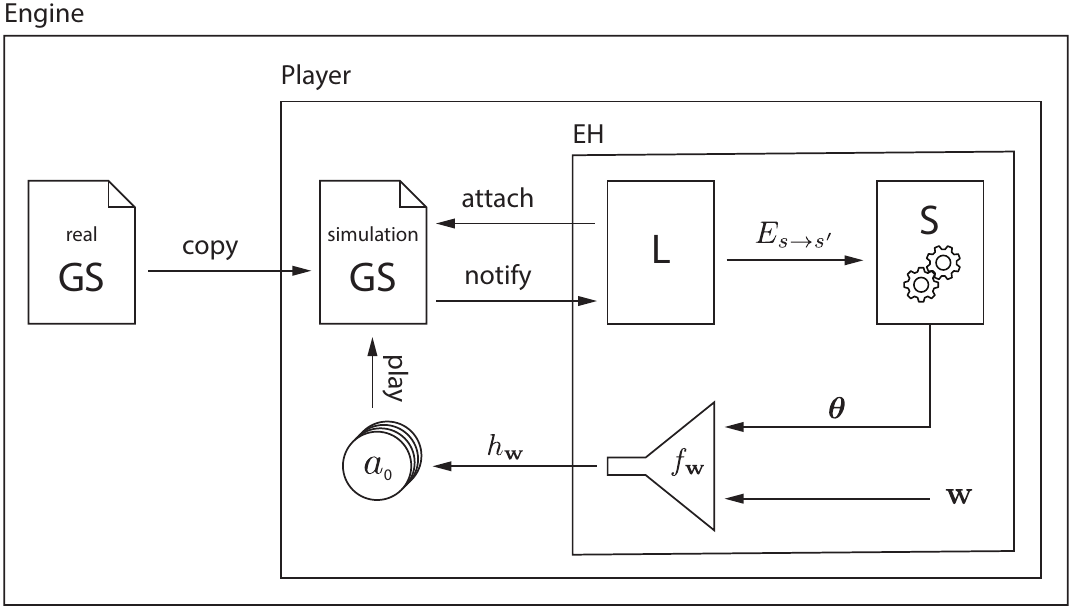}
		    \caption{\label{fig:loggingflow} Interaction between engine, state, player and EF. The player starts simulating the action sequence from action a$_0$. At the end of the evaluation the EF can be queried to get the value $v$.}
		\end{figure}

 		An EF contains two components: a synthesis function $\sigma$ (S), an event logger (L) and a set of weights $w$. The Event Logger receives all the events triggered by the game state (GS) it is attached to. Then, when the action sequence is evaluated, the events are forwarded to the synthesizer. S is responsible for filtering and processing the events in order to produce a vector of features $\boldsymbol\theta$ of the same length of $\mathbf{w}$.
		As the EF-based player prepares to evaluate an action sequence, the EF is attached to the game state used for the forward planning, then the actions are performed. At this point, the EF can compute the value $v=h_{\mathbf{w}}(\boldsymbol\theta$). See Figure \ref{fig:loggingflow} for a visual representation.

	\subsection{Synthesis Function}

		Our synthesis function $\sigma$ is quite straightforward: it counts the events grouping them by \textit{type} filtering out the events not triggered by the player.
		In order to reduce the size of the feature vector we also introduce the possibility of remapping the types to group them. See Table \ref{table:events} for the mappings, $id$ is the identity mapping while $hc$ is a hand-crafted mapping. Note that mapping to $-1$ is a way of discarding the event altogether.

    \subsection{Mixer Function}

		In an effort to reducing the complexity of this preliminary study we use two basic models: \textit{linear} function consisting in simply multiplying each features by a weight, i.e. $lin_{\mathbf{w}}(\boldsymbol\theta)= \mathbf{w}\bullet\boldsymbol\theta$; \textit{multivariate polynomial} function of degree $d$, see Equation \ref{eq:poly}. The function $sel$, selects the $j$-th element in the $i$-th $d$-multicombination of $| \boldsymbol\theta |$ variables.

		\begin{equation}
			\label{eq:poly}
			poly^{d}_{\mathbf{w}}(\boldsymbol\theta) = \sum_{i=1}^{\multiset{|\boldsymbol\theta|}{p}} w_{i}\prod_{j=1}^{d}sel_{\boldsymbol\theta}(i,j)
		\end{equation}

		For example given $\boldsymbol\theta = [\theta_0,\theta_1]$ we have $poly_{\mathbf{w}}^{2}(\boldsymbol\theta)=w_1\theta_0^2+w_2\theta_0\theta_1+w_3\theta_1^2$. We want to emphasise that we don't need any constant $w_0$ since $h_{\mathbf{w}}$ is a ranking function. The purpose of using a higher degree function is to detect possible dependencies between features.

		The space of possible weights $\mathbf{w}$ then becomes the hyperparameter space for the mixer function $f_{\mathbf{w}}(\boldsymbol\theta)$;

		\begin{table}[]
		\caption{Table showing the number of weights for each mixer function given two different event mappings.}
		\label{table:weights}
		\resizebox{\columnwidth}{!}{
		\renewcommand*{\arraystretch}{1.4}
		\begin{tabular}{cc}
		\begin{tabular}{l|l|l}
		\textbf{Mixer} & \textbf{Features}& \textbf{Weights}\\
		\hline
		$lin_{\mathbf{w}}^{hc}$    & 5    & 5               \\
		$poly_{\mathbf{w}}^{2,hc}$ & 5    & 15              \\
		$poly_{\mathbf{w}}^{3,hc}$ & 5    & 35              
		\end{tabular} &
		\begin{tabular}{l|l|l}
		\textbf{Mixer}             & \textbf{Features}& \textbf{Weights}\\
		\hline
		$lin_{\mathbf{w}}^{id}$    & 18               & 18              \\
		$poly_{\mathbf{w}}^{2,id}$ & 18               & 171             \\
		$poly_{\mathbf{w}}^{3,id}$ & 18               & 1140		
		\end{tabular}
		\end{tabular}}
		\end{table}

		\subsection{The implementation burden}
			All these functionalities come expensive, creating and embedding into a game a logging infrastructure together with mixer and synthesising functions requires time and engineering skills. However we need to make two observations to put things into perspective.
			
			First, this approach has been developed with the long term objective for being applied in a playtesting scenario where there is an explicit need for expressing as many strategies as possible in the most controlled way. Take the example of an agent based on an Artificial Neural Network (ANN) that given the current state as input provides next action to play. The control we have over such agent, and consequently its strategy, is by adjusting the ANN's weights, unfortunately selecting the weights is a very delicate. In the EF instead there's a direct link between game dynamics (represented as features) and the agent's behaviour.

			Second, what is currently done by hand could be done by a specialised model trained using classic RL techniques, such model would receive as input the starting and ending states and output the features vector. There's a trend that is moving from single network architectures to multi-network architectures, i.e. from the Deep Neural Networks in \cite{mnih2013playing} to AlphaStar \cite{vinyals2019alphastar}. This forces each portion of the system to focus and specialise into a specific task. What we are envisioning is the possibility in the near future to automatically generate the features $\boldsymbol\theta$ by learning the mapping we are currently hand-crafting in $\sigma(E_{s \to s'})$ with the aid of the logging system.

\section{Experiments} \label{sec:experiments}

	The main objective of the experiments is to show that event-value functions can be a substitute for using score variation SFP algorithms.
	We designed three sets of experiments:
	\begin{itemize}
		\item BMRH+EF Tuning: hyperparameter tuning using NTBEA (see Section \ref{sec:background:ntbea}) in the hyperparameter space of BMRH coupled with the hyperparameter space of a EF, run for several EFs (see Section \ref{sec:experiments:ntbea});
		\item Round-robin Tournament: picking the best agents tuned with NTBEA we are going to set one against the other (see Section \ref{sec:experiments:roundrobin});
		\item Multi-opponent Games: compare the stability in terms of win rate in scenarios where there 3 opponents (see Section \ref{sec:experiments:multiopponent}).
	\end{itemize}
	All the SFP agents in the following experiments were allowed a budget of 1000 simulated actions for each turn of the agent.

	\subsection{BMRH+EF Tuning} \label{sec:experiments:ntbea}

		\begin{figure*}[!t]
	    \centering
	    \includegraphics[width=\textwidth]{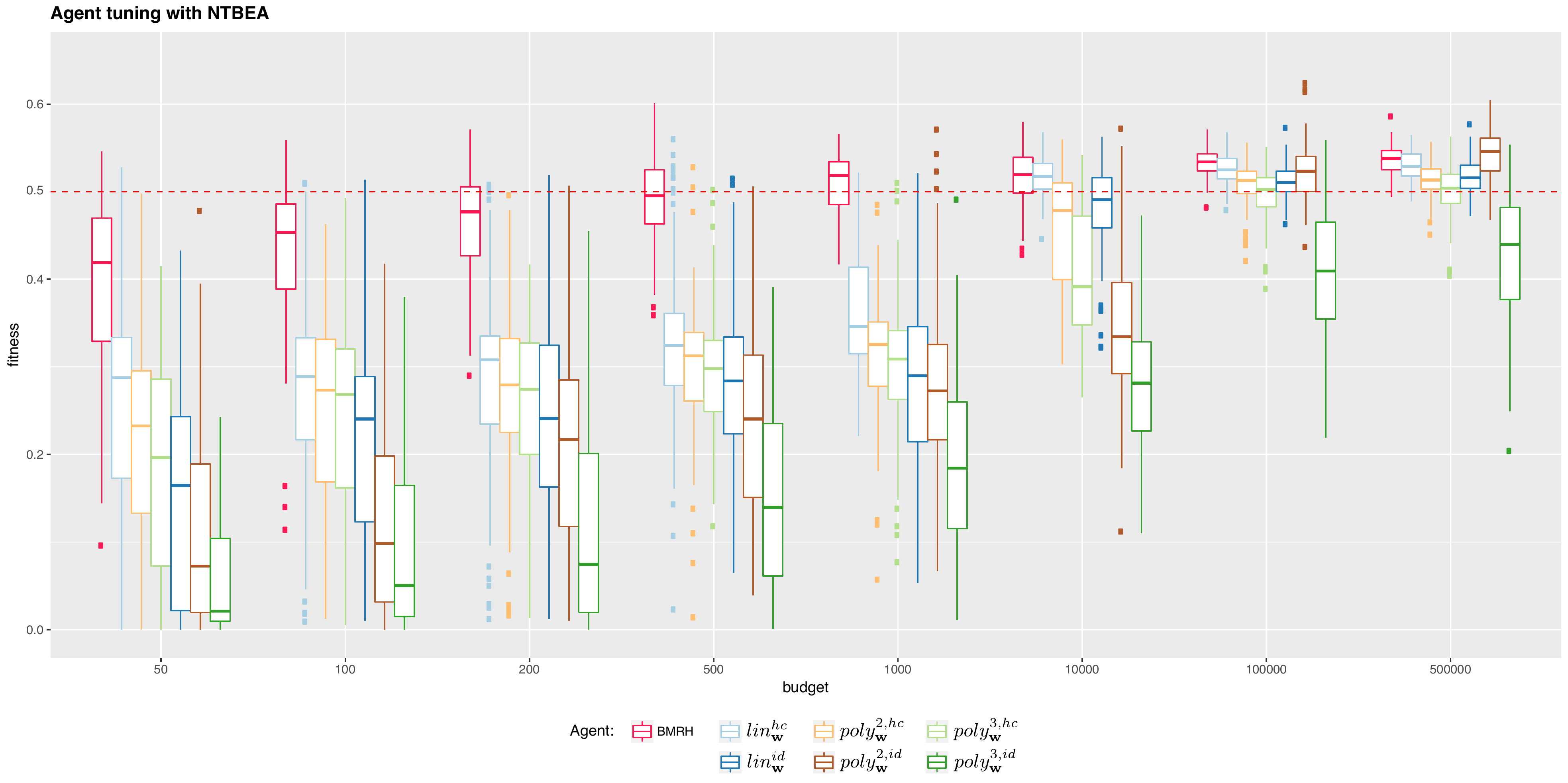}
	    \caption{\label{fig:ntbea} NTBEA results all the configurations of BMRH+EF as the budget variables.}
	    \end{figure*}

		The event-value function is parameterised in a set of weights $\mathbf{w}$, these define an hyperparameter space of $|\mathbf{w}|$ dimensions. Since NTBEA is a discrete optimisation algorithm, we discretise the continuous space of each weight $w_i$ as $\mathbf{W} = \{-1, -0.8, -0.6, -0.4, -0.2, 0, 0.2, 0.4, 0.6, 0.8, 1\}$. The limited variety of weights could constrain the ability of fine tuning the heuristics, but it will most likely be sufficient for the lower dimensional $\mathbf{w}$.
		
		When BMHR uses an EF, BMHR's and EF's hyperparameter spaces are combined to define the hyperparameter space of BMRH+EF, combining the two spaces will allow the two components to adapt to each other.
		We are going to tune several configurations of BMRH+EF to perform as well as possible against BMRH*, therefore NTBEA will be set to maximise the win rate of BMRH+EF in a 2-player Splendor game. For each set of hyperparameters evaluated, we run a single game in spite of the high stochasticity of \gameR, in fact, NTBEA was designed to deal with objective functions with high noise.
		
		In Table \ref{table:weights} are reported all the mixer functions used in the experiments, when considering the dimensionality of the hyperparameter space for those experiments we need to add 10 dimensions from BMRH to the dimensions specified in the table.
		NTBEA was run 100 times for each BMRH+EF configuration and with several budgets: 50, 100, 200, 500, 1k, 10k, 100k, 500k games. Once the budget is over, the suggested optimal configuration is validated on 1000 games against BMRH*. Under the same conditions we also tuned the BMRH, but with the classic score-based value function, as this will give us an idea of the trade off between tuning a basic agent with a small hyperparameter space (10D) and a more sophisticated agent with a much larger space (from 15D to 1150D).
		NTBEA has two parameter that can be adjusted, we picked the values $k=1$ and $\epsilon=0.7$, see \cite{lucas2018n} for more details.
		The results from the validation phase after NTBEA are shown in Figure \ref{fig:ntbea}.

	\subsection{Round-robin Tournament} \label{sec:experiments:roundrobin}
		
		The main purpose of running a round-robin tournament is to get a better understanding of the all-round performance of the optimised agents, not only against the reference agent BMRH*. From the previous NTBEA experiments we selected the best (highest win rate) configuration according to the validation statistics, we are going to name this configurations by using the pedice $\mathbf{w}^{*}$ instead of the generic $\mathbf{w}$ (e.g. $poly_{\mathbf{w}^{*}}^{3,hc}$). Instead BMRH** is the optimal BMRH tuned against BMRH*.
		In Figure \ref{fig:roundrobin} the win percentage of each couple of agents is reported based on 10000 games of two-player Splendor. This number of samples guarantees that the real value of the estimated win percentage will lie within a 95\% CI with boundaries $\pm 1$.

	\subsection{Multi-opponent Games}\label{sec:experiments:multiopponent}
		Since one of the higher sources of stochasticity in \gameR~are the opponents, we tried to evaluate the robustness of the agents by varying the number of opponents.
		We first let the agents play against 3 RND agents in a 4-player Splendor game for 1000 times. This is to evaluate if the presence of a random player can influence the their performance. Then we repeat the experiments but with 3 BMRH* opponents.
		Figure \ref{fig:4players} shows the outcome of these experiments.

\section{Discussion} \label{sec:discussion}

	\subsection{BMRH+EF Tuning} \label{sec:discussion:ntbea}

		The tuning experiments have highlighted a multifaceted scenario concerning both the optimisation algorithm and the BMRH+EF spaces. The results reported in Figure \ref{fig:ntbea} show the box plot summarising the 100 runs for each experimental condition, outliers have been reported with single dots, as usual.
		Generally we can see that the average quality of the tuned agents increase as the budget does. Not only the performance improves but the whiskers also shrink denoting more consistent convergence of the tuning algorithm. Both are to be expected and they highlight the sanity of the experiments.

		\subsubsection{Overall considerations}

			First of all, let's discuss our baseline, the BMRH hyperparameter space, on average it can be tuned to outperform BMRH* using a budget of 1000 games, this is a very small amount of computational resources especially for such a fast framework. As the budget increases the performance saturates never crossing the 60\% mark.
			Looking at the performance of BMRH we can notice how it finds it's best configuration with a small budget of 500. This suggests that NTBEA could probably make a better use of the budget pushing for more exploration, or this could be symptomatic of a limit in dealing with big amounts of data in small search spaces.
			
			With regards to the BMRH+EF spaces, in order to find well performing solutions, NTBEA needed a budget of 10k games for $lin_{\mathbf{w}}^{hc}$, this was expected as the search space has 5 more dimensions. It is actually reassuring as it took only 10 times the budget (w.r.t. BMRH) for a search space $>10^5$ times bigger.
			The best overall performance, with a win rate of 62\%, was found using the $poly_{\mathbf{w}}^{2,id}$ EF and a budget of 100k. This is an outlier considering the average outcome for such experimental condition. However the purpose of these experiments is not evaluating NTBEA but rather the possibility for EFs to improve BMRH performance, thus outliers are just as relevant as any other data point.
			A budget of 500k is enough to find an average tuning above the 50\% threshold with the exception of the two $poly_{\mathbf{w}}^{3}$ .
			
			Beyond the problematic of dealing with more weights to tune we can see that actually providing a richer set of features can yield better peak performance. This is the case for $poly_{\mathbf{w}}^{2,id}$ that can find better tuning than $poly_{\mathbf{w}}^{2,hc}$ in spite of a much bigger dimensionality: 15 vs 171.

			We also want to emphasise that with the current approach, even though we can recognise score variations with events 16 and 17, we are not actually using their magnitude. So EFs are not able to differentiate between a variation of 3 points and that of 1. The same is true for the amounts of coins taken or given. This is an area of improvement that will require to query information stored in other fields of the event data structure. In this work, however, we stuck to the simple use of \textit{type} and \textit{who} fields of an event, for the sake of simplicity.

		\subsubsection{Optimal configurations}
			One of the most interesting aspects of these new EFs is how they can be used to better understand the game and the player's strategy.
			Both $ lin_{\mathbf{w}^{*}}$ can be used to directly infer the relevance (or the agent's preference) of the events from \textit{hc} and \textit{id}.

			In the former case, $ lin_{\mathbf{w}^{*}}^{hc}$, the optimal weights found are (0.2, 0.2, -0.4, -0.6, 0.8), matching these with Table \ref{table:events} we can say the following: taking tokens is important and even more important if different token types are taken as they trigger more events; events bringing points to the player are the most important; reserving cards is mildly discouraged unless if hidden which is lightly encouraged; attracting nobles is seen as a negative event $w_3=-0.6$, however when considering that this event is always triggered together with the event linked to $w_4=0.8$ it becomes apparent that this is just a way of preferring point events that come from cards. This shows how this new approach is able to differentiate between action sequences leading to the same score variation.
			
			In the latter case, $ lin_{\mathbf{w}^{*}}^{id}$, the optimal configuration was (-0.8, 0.2, -0.4, -1.0, 0.8, 0.2, -0.2, -0.2, 0.8, -0.2, -1.0, -0.8, -0.8, 0.2, 1.0, 0.8, 1.0, 0.4). In this configuration we can see the same preference of card points over noble points expressed by the previous case. This time, however, it is explicitly stated in $w_{16}=1.0$ and $w_{17}=0.4$. Another interesting insight is given by $w_{9}=-0.2$ and $w_{11}=-0.8$ showing how common tokens are less valuable than joker tokens. 

			These results are the demonstration that using an EF the agent can express very specific and refined strategies in terms of in-game behaviour. Unfortunately interpreting the results for the $poly_{\mathbf{w}^{*}}$ EFs can be very complicated, the optimal configurations can be found in the online repository\footnote{at the path \textit{/agents/R2-NTBEA-best.json}}.

			Looking at the hyperparameters of the agent is also important as these are tuned as well. We saw from \cite{bravi2019rinascimento} that all the agents were tuned with a sequence length of 2 actions. This time, instead, several of the optimal configurations had a sequence length of three actions, namely: BMRH**, $poly_{\mathbf{w}^{*}}^{2,hc}$ and $poly_{\mathbf{w}^{*}}^{3,hc}$. The others instead were tuned with a length of 2, except $poly_{\mathbf{w}^{*}}^{3,id}$ with length 1. These longer horizons highlight the need for longer planning in order to beat BMRH*.

			\begin{figure}[!t]
		    \centering
		    \includegraphics[width=\columnwidth]{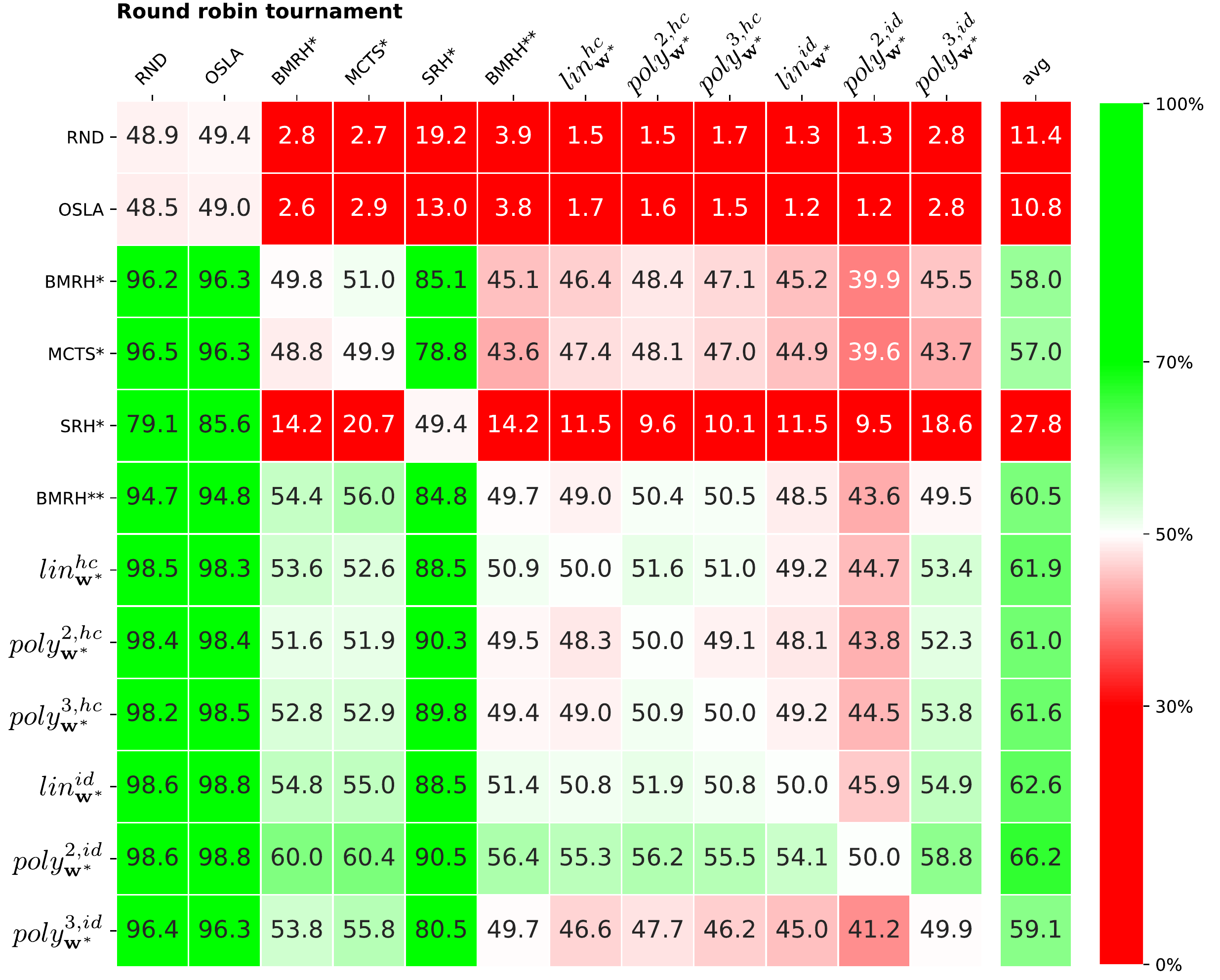}
		    \caption{\label{fig:roundrobin} Results for the round-robin tournament, each element $(row, column)$ shows the win rate of player $row$ against player $column$ . The heatmap colours are invariant in 0-30\% and 70-100\% in order to highlight relevant subtleties around the 50\% mark. The last column, \textit{avg}, shows the average win rate across all opponents.}
			\end{figure}


	\subsection{Round-robin Tournament}\label{sec:discussion:roundrobin}
		At first glance at the results reported in Figure \ref{fig:roundrobin}, we can see how the top section of the map tends more towards red tones, these are the non-EF agents. This is also suggested by a higher average performance across opponents (reported on the \textit{avg} column). Our baseline is BMRH** shows an improved tuning compared to BMRH*, as they share the same hyperparameter space. However when playing against an EF-BMRH agent, it shows a 50/50 win rate at best. The result that immediately stands out is the performance of $poly_{\mathbf{w}^{*}}^{2,id}$, this agent dominates all the other agents, in particular BMRH** with a solid 56.4\% of victories. This result is encouraging as it shows a promising edge over a points-based agent.

		When it comes to comparing different EFs, the differences becomes more subtle with the exception of $poly_{\mathbf{w}^{*}}^{3,id}$ and $poly_{\mathbf{w}^{*}}^{2,id}$ . The first is simply the weakest of all EFs, the second, instead, seems to dominate all the others. This highlights the potential of catching dependencies between features.

		What is striking, looking at these results and at NTBEA's, is the lack of a really strong player with win rates around 80/90\%.
		There is a number of reasons for this. The most likely reason is the heavy interference of stochasticity in the game which can suddenly turn the tables. Another reason is the limited amount of budget allowed to the agents (i.e. 1000 simulated actions per turn). This is a limiting factor in their ability to predict the future, however, the amount of stochasticity could also just be prohibitive. Finally, we must remind that the real search space for the weights is continuous, thus the limited set of weights used to discretise such space might be limiting the NTBEA's ability of fine tuning the functions with higher number of weights.

		Finally, $poly_{\mathbf{w}^{*}}^{2,id}$ shows how catching relationships between the events can be important. This is more precise as more the information is detailed, in fact, it uses the complete set of events. Its search space is quite big, 171 dimensions, but evidently enough to be properly search within the budgets allowed.

	\subsection{Multi-opponent Games}\label{sec:discussion:multiopponent}
\begin{figure}[!t]
	    \centering
	    \includegraphics[width=\columnwidth]{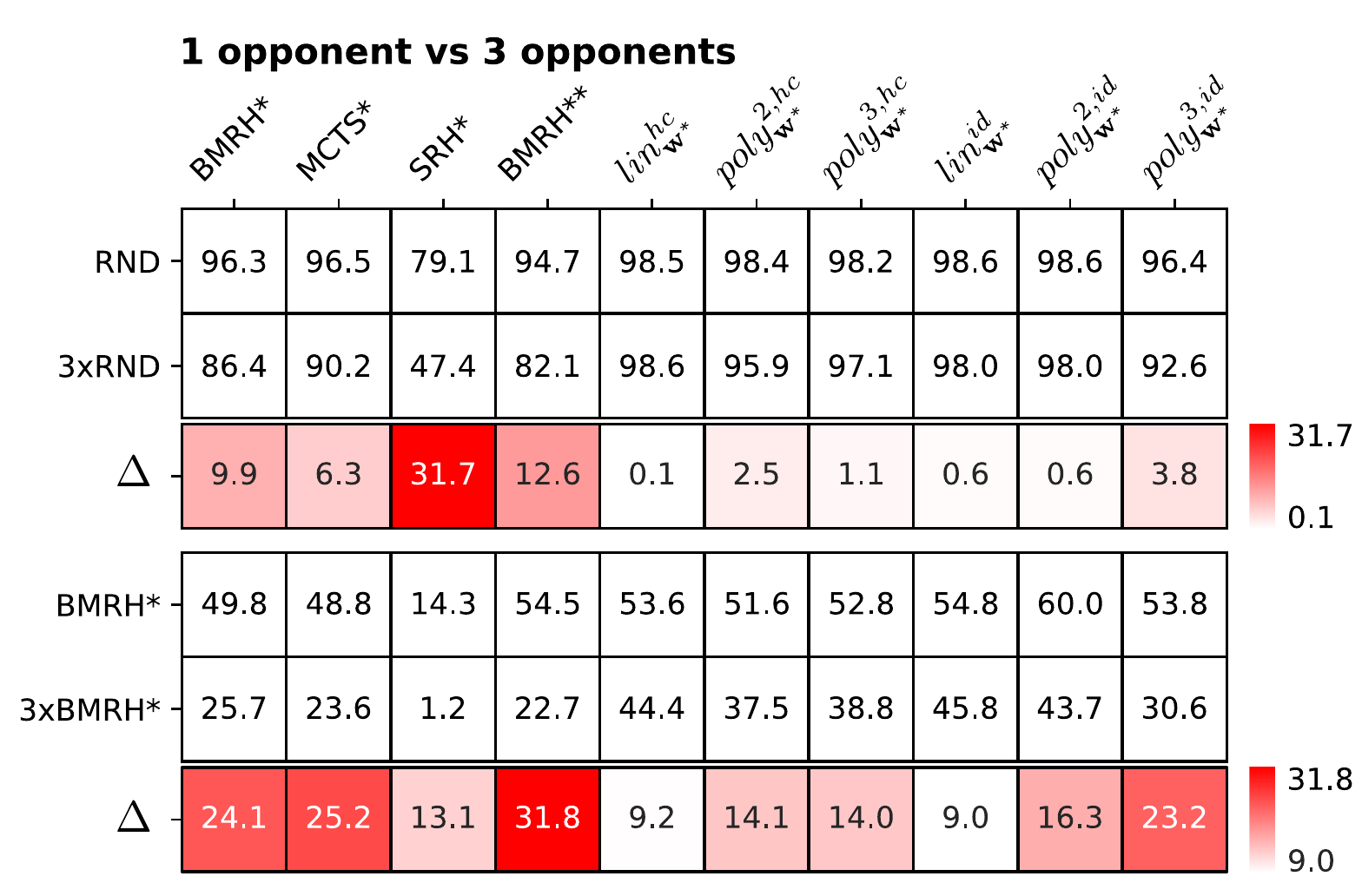}
	    \caption{\label{fig:4players} Win rates for the column players against the row agents. In each one of the two matrices, the third row represents the delta between the win\%. For a 4-player game the uniform random target for the win rate is 25\%.}
		\end{figure}

		In the first scenario where the agents were tested against three RND agents we can see some minor performance drops but they become significant as we jump to the games with the three BMRH*, see Figure \ref{fig:4players}. This is true for all agents, however the agents not using EFs show the most noticeable drops. From the \textit{delta} rows in Figure \ref{fig:4players} we can see that: in the case of RND opponents, EFs' biggest drop is 3.8\% while the smallest for non-EF is 6.3\%, a significant margin; instead, in the case of BMRH* opponents, EF biggest drops are still moderate compared to non-EF that can go as bad as a whopping 31.8\%.
		These results show how on average the EF based agents can guarantee a much more consistent performance in more noisy and competitive games.
	 
\section{Conclusions and Future work} \label{sec:conclusion}
	
	In this paper we have presented a novel approach for value functions in scenarios with scarce or absent reward signals: the event-value function $h_{\mathbf{w}}(E_{s \to s'})$. The main purpose is to provide a smoother gradient in the evaluation of a sequence of actions. This evaluation will be based on the set of events that are triggered during the execution of such actions. These events are characterised by a type. Discerning by type, we can count the number of events that happened and create a feature vector. This feature vector describes the dynamics of the system uncovering what happens in the state transitions. Finally by recombining and weighting the features we can evaluate the quality of the dynamics triggered by the action sequence.

	This novel approach has shown performance improvements in terms of win rate when compared to the baseline agents BMRH* and BMRH**. In fact, these baselines can obtain the same win rates at best (see Section \ref{sec:discussion:roundrobin}).
	Event-value functions have also proven more robust when playing against more opponents, as shown in Section \ref{sec:discussion:multiopponent}. EFs also allowed a more controllable characterisation of the game-playing style of the agent as shown by the EF's weights analysis. Using an event-value function we can discern the difference between two action sequences that lead to the same outcome in terms of score variation, meaning we have finer control over the behaviour of the agent. This particular feature is very important when it comes to the ability of automatically play-testing a game, in fact optimal agents are blind to the variety of possible non-optimal but human-like strategies in the game.

	For future work we plan on doing a variety of improvements and enhancements to the work presented here and also to apply this to different scenario.
	
	The first improvement could be using more sophisticated mixing functions using models that can express more sophisticated functions than the ones explored here. The use of ANNs seems appropriate as they could be able to detect more dependencies between features. However, this approach will require the use of a optimisation algorithm for continuous spaces, since tuning the weights of a ANN using a discretised space can be much less productive than a linear function.
	
	Another improvement could be brought in the use of the information from the event, this far, only the \textit{type} and \textit{who} were used. Embedding the richer information coming from the event's \textit{attributes} can potentially allow for a more precise definition of the player's strategy. An option could be assigning weights to these values as well while carefully distinguish between categorical, ordinal and numerical attributes.

	Since EFs could define and express more focused strategies, several behavioural metrics could be defined and used to numerically evaluate the differences between the agents presented so far. A more qualitative and descriptive approach to developing game-playing AI is probably the way forward for more believable agent as well.

	Finally, the hyperparameter space of the EF-based agents could be explored and, using several metrics as the ones just mentioned, the strategic space of a game could be outlined and the most promising behavioural strategies highlighted. This process could be approached using the MAP-Elites algorithm that has shown very promising results \cite{mouret2015illuminating}.



\section*{Acknowledgements}
	\addcontentsline{toc}{section}{Acknowledgements}
	This work was funded by the EPSRC CDT in Intelligent Games and Game Intelligence (IGGI) EP/L015846/1.
	This research utilised Queen Mary's Apocrita HPC facility \cite{king_thomas_2017_438045}, supported by QMUL Research-IT.
	Special thanks to Valerio Bonometti for the lengthy discussions, confrontation and rubber duck debugging.

\bibliographystyle{IEEEtran}
\bibliography{bib}

\end{document}